# Natural Scene Recognition Based on Superpixels and Deep Boltzmann Machines

Jinfu Yang[1], Jingyu Gao[1*], Guanghui Wang[2], Shanshan Zhang[1]

[1] Department of Control Science & Engineering, Beijing University of Technology, NO.100 Chaoyang district, Beijing, 100124, P.R.China.

[2] Department of Electrical Engineering & Computer Science, University of Kansas, Lawrence, KS 66045-7608, USA

* Corresponding author

E-mail: gjy890316@126.com (JG)

# Abstract

The Deep Boltzmann Machines (DBM) is a state-of-the-art unsupervised learning model, which has been successfully applied to handwritten digit recognition and, as well as object recognition. However, the DBM is limited in scene recognition due to the fact that natural scene images are usually very large. In this paper, an efficient scene recognition approach is proposed based on superpixels and the DBMs. First, a simple linear iterative clustering (SLIC) algorithm is employed to generate superpixels of input images, where each superpixel is regarded as an input of a learning model. Then, a two-layer DBM model is constructed by stacking two restricted Boltzmann machines (RBMs), and a greedy layer-wise algorithm is applied to train the DBM model. Finally, a softmax regression is utilized to categorize scene images. The proposed technique can effectively reduce the computational complexity and enhance the performance for large natural image recognition. The approach is verified and evaluated by extensive experiments, including the fifteen-scene categories dataset the UIUC eight-sports dataset, and the SIFT flow dataset, are used to evaluate the proposed method. The experimental results show that the proposed approach outperforms other state-of-the-art methods in terms of recognition rate.

## Introduction

Scene recognition, also called scene classification, is a necessary procedure of the humans' vision system. As an important topic of computer vision, scene recognition is used to accurately and immediately observe the surrounding environments; and has attracted more and more attention due to the potential wide applications such as automatic driving and robot navigation.

Most early work in scene recognition focuses on extracting surface features to recognize scenes or objects. For example, color and texture are usually considered to be surface features and are widely used to represent features of scene images. *Haralick et al.* [1] employed the features of texture, color and frequency to infer high-level information for indoor and outdoor categorization. However, the features mentioned above are neither accurate nor sufficient enough to present the information of images because they are unrepresentative and indistinguishable. For instance, the color blue can represent either the sea or sky in scene images.

In recent years, many researchers have paid attention to deep-seated features rather than surface features. *Zheng et al.*[2] proposed a mid-level image representation, called Hybrid-Parts, which was generated by pooling the response maps of object part filters, to represent compact information of input images. *Jiang et al.* [3] presented a novel image representation method, called Randomized Spatial Partition (RSP), which was characterized by the randomized partition patterns. The method makes it possible to extract the most descriptive layout features for each category of scenes. *Sadeghi and Tappen*[4] introduced a latent pyramidal region (LPR) method . They used latent SVM framework as region detectors to capture the key characteristics of the scenes. *Lin et al.*[5] proposed a joint model for scene classification, which used part appearance and important spatial pooling regions (ISPRs) to reduce the influence of false responses. In addition, they illustrated the promising results by combining the ISPR with an improved fisher vector (IFV) algorithm. However, the features mentioned above are incapable of representing the hidden information of images well, as the local features are always defined manually.

Recently, global features have attracted more attention from researchers again. *Hinton* and *Salakhutdinov* [6] proposed a multiple restricted Boltzman machine, named deep belief networks (DBN), and a greedy layer-wise learning algorithm. They introduced an effective way of initializing weights that allows deep networks to learn low-dimensional information, which works much better than the principal error of back propagation in reducing the dimensionality of data. Compared to other algorithms, it is an unsupervised learning method which makes it more competent for big data and has been succesfully applied in hand written digit recognition[7], object recognition[8], and speech recognition[9]. In 2012, *Salakhutdinov* and *Hinton*[10] proposed another model of deep learning, called deep Boltzman machines (DBM), by changing the structure of the DBN model; and the model enables the neighbouring layers to represent each other. As a result, the visible layer can receive back-propagation error of output layer and the model can fine-tune its parameters in a layer-by-layer way to decrease the errors. The DBM obtains better performance than other deep learning algorithms in object recognition, hand written digits recognition [10][11], and multimodal containing both word and image learning[12].

However, when the DBM is used to extract features for natural scene recognition, the issue of computational complexity must be considered, since it requires a lot of matrix operations and hundreds of interactions with large images as input data. Convolution and pooling are suggested to work with large images[13], however, the convolution has high computational complexity, and pooling is limited by the coordinates of pixels. In this paper, we propose a new natural scene recognition method based on superpixels and the DBM. First, large-sized natural scene images are preprocessed by the SLIC algorithm to reduce their sizes. Superpixels are generated through grouping the pixels into atomic regions where the pixels are assigned to the same labels defined by the distance between the cluster center and each pixel in the given region. Then, a two-layer DBM model is constructed by stacking two RBMs (see Fig. 3(a)), and the superpixels are regarded as the input data to train the DBM model in a layer-by-layer way. After training the first two layers of the DBM model (i.e., the first RBM), the parameters are frozen and the second RBM is trained by using a greedy layer-wise algorithm, with the output of the first RBM used as the input of the second RBM. Other layers of the DBM model are trained in the same way. Finally, a softmax classifier is employed to classify the extracted features. The main

contributions of this paper are as follows:

(1) We propose a novel scene recognition method that combines the superpixels and the DBM for categorizing large-sized natural images.

(2) The superpixels-based preprocessing strategy can obtain better performance than the convolution and pooling for the DBM in recognizing natural scene images.

(3) The proposed method performs better than other counterparts in terms of recognition rate.

The remainder of this paper is organized as follows. In section 2, the structure of the restricted Boltzmann machine and its learning procedure is introduced. Section 3 presents the architecture of the DBM and the greedy layer-wise algorithm for learning the DBM model. Section 4 describes the process of image preprocessing with the SLIC. The proposed scene recognition method with the softmax regression is elaborated in Section 5. The performance of the proposed method is evaluated and discussed in Section 6 using three datasets. Finally, the paper is concluded in Section 7.

## Background on RBM and Model Learning

In this section, we introduce the RBM, which is the basis of the DBM. Since the DBM is a stacked RBMs, the training process of each DBM layer is the same as that of the RBM.

### Restricted Boltzmann Machine

The RBM is the variant of Boltzmann machine[14], and it consists of visible units $v \in \{0,1\}^p$ and hidden units $h \in \{0,1\}^p$. As shown in Fig. 1, the units in neighboring layers are connected by the weights, but the units in the same layer do not link with each other in order to reduce the computational redundancy.

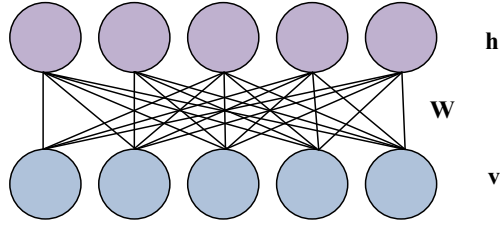

**Fig. 1 Interaction graph of an RBM.** $\mathbf{v}$ represents visible units and $\mathbf{h}$ denotes hidden units. $\mathbf{W}$ represents the weight matrix between the visible units and the hidden units, where the number of rows and columns of $\mathbf{W}$ equals to the number of the visible units and hidden units.

The probability parameter of the RBM is $\theta(\mathbf{W},\mathbf{b},\mathbf{c})$, where $\mathbf{W}$ represents the weight matrix between the visible layer $\mathbf{v}$ and the hidden layer $\mathbf{h}$; and $\mathbf{b}$ and $\mathbf{c}$ denote hidden and visible unit biases, respectively. The joint energy function of an RBM is defined as

$$E(\mathbf{v},\mathbf{h};\boldsymbol{\theta}) = -\sum_{i,j} v_i W_{i,j} h_j - \sum_i b_i v_i - \sum_j c_j h_j \qquad (1)$$

where $i$ and $j$ are respectively the iterations of the visible and hidden units. From equation (1), a good model of data with the parameter $\boldsymbol{\theta}$ is interpreted as a model that has high energy in regions of low data density and low energy elsewhere [15]. The matrix $\mathbf{W}$ of connection weights between units is symmetric.

According to the joint energy $E(\mathbf{v},\mathbf{h};\boldsymbol{\theta})$, an RBM associates to each state $v \in \{0,1\}^p$ and $h \in \{0,1\}^p$, and the probability is given by

$$p(\mathbf{v},\mathbf{h}) = \frac{1}{Z(\boldsymbol{\theta})} \sum_{\mathbf{h}} \exp(-E(\mathbf{v},\mathbf{h},\boldsymbol{\theta})) \qquad (2)$$

where $\exp(\cdot)$ is the natural exponential function, *and* $Z(\boldsymbol{\theta})$ is a normalization constant given by

$$Z(\boldsymbol{\theta}) = \sum_{\mathbf{v}} \sum_{\mathbf{h}} \exp(-E(\mathbf{v}, \mathbf{h}, \boldsymbol{\theta})) \tag{3}$$

From the above equations, the conditional probability of each unit in one layer being activated given the units in the other layer is defined as

$$p(\mathbf{h}|\mathbf{v}) = \prod_j p(h_j|\mathbf{v}) \quad p(h_j = 1|\mathbf{v}) = \sigma\left(-\sum_i W_{ij} v_i - c_j\right) \tag{4}$$

$$p(\mathbf{v}|\mathbf{h}) = \prod_j p(v_i|\mathbf{h}) \quad p(v_i = 1|\mathbf{h}) = \sigma\left(-\sum_i W'_{ij} h_j - b_i\right) \tag{5}$$

where $\sigma(\bullet)$ represents the sigmoid function as $\sigma(x) = \dfrac{1}{1+e^{-x}}$.

## Model learning

Maximum likelihood estimation (MLE) is a typical approach to learning the parameters of the RBM. The gradient ascent of the log-likelihood with respect to $\boldsymbol{\theta}$ is

$$\frac{\partial}{\partial \boldsymbol{\theta}} L(\boldsymbol{\theta}) = -\left\langle \frac{\partial E(\mathbf{v}; \boldsymbol{\theta})}{\partial \boldsymbol{\theta}} \right\rangle_{data} + \left\langle \frac{\partial E(\mathbf{v}; \boldsymbol{\theta})}{\partial \boldsymbol{\theta}} \right\rangle_{model} \tag{6}$$

where $E(\mathbf{v}; \boldsymbol{\theta})$ is given by equation (1), and $\langle \bullet \rangle_{data}$ represents the expectation of all visible vectors $\mathbf{v}$ in regard to the data distribution, $\langle \bullet \rangle_{model}$ denotes the model distribution defined by equation (2). Unfortunately, computing expectation involves an exponential number of terms, which makes it difficult. Therefore, the contrastive divergence (CD) [16] is commonly adapted to learn the RBM model by maximizing the likelihood. Based on the CD algorithm, the parameters are updated as below.

$$\begin{aligned} W_{ij} &= W_{ij} + \eta_1 \left(v_i^0 h_j^0 - v_i^n h_j^n\right) \\ b_i &= b_i + \eta_2 \left(v_i^0 - v_i^n\right) \\ c_j &= c_j + \eta_3 \left(h_j^0 - h_j^n\right) \end{aligned} \tag{7}$$

where $\eta_i (i=1,2,3)$ is the learning rate for each parameter, $\mathbf{v}^0$ takes the value from the observed data distribution, $\mathbf{h}^0$ is defined by equation (4), $\mathbf{v}^n$ is acquired from the sampled data by $n$-step Gibbs sampling, and $\mathbf{h}^n$ is obtained from equation (4) based on $\mathbf{v}^n$.

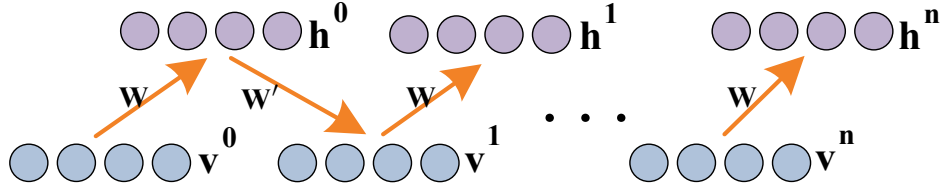

**Fig. 2 The procedure of Gibbs sampling.** Given $\mathbf{v}$, $p(\mathbf{h}|\mathbf{v})$ is computed. Given $\mathbf{h}$, $p(\mathbf{v}|\mathbf{h})$ is computed. Repeat this procedure $n$ times, and obtain the values of $\mathbf{v}^n$ and $\mathbf{h}^n$.

We obtain $\mathbf{v}^n$ and $\mathbf{h}^n$ using the alternative Gibbs sampling, as shown in Fig. 2. During CD learning, Gibbs sampling [17] is initialized at each variable, and it only requires a few steps to approximate the model distribution [10]. To introduce the procedure of Gibbs sampling, we define $h_j^n$, where $n$ denotes the *n-th* Gibbs sampling and $j$ denotes the number of units. First, for all hidden units $h_i^0$, compute $p(h_i^0|\mathbf{v}^0)$ where $\mathbf{v}^0$ stands for the original visible units; $h_i^0 \in \{0,1\}$ equals to $p(h_i^0|\mathbf{v}^0)$. Then, compute $p(v_j^1=1|\mathbf{h}^0)$ and let $\mathbf{v}^1$ equals to $p(v_j^1=1|\mathbf{h}^0)$, where $\mathbf{v}^1$ stands for the reconstructed data of the input data. Next, compute $p(h_i^1=1|\mathbf{v}^1)$ and the value of $\mathbf{h}^1$ is replaced by $p(h_i^1=1|\mathbf{v}^1)$ where $\mathbf{h}^1$ stands for the reconstruction data of $\mathbf{h}^0$. Repeat the above steps $n$ times to obtain $\mathbf{v}^n$ and $\mathbf{h}^n$.

## Deep Boltzmann Machines

A DBM is stacked by the RBMs. Unlike the DBN model [7], the DBM combines bottom-up and top-down passes and has better generative property.

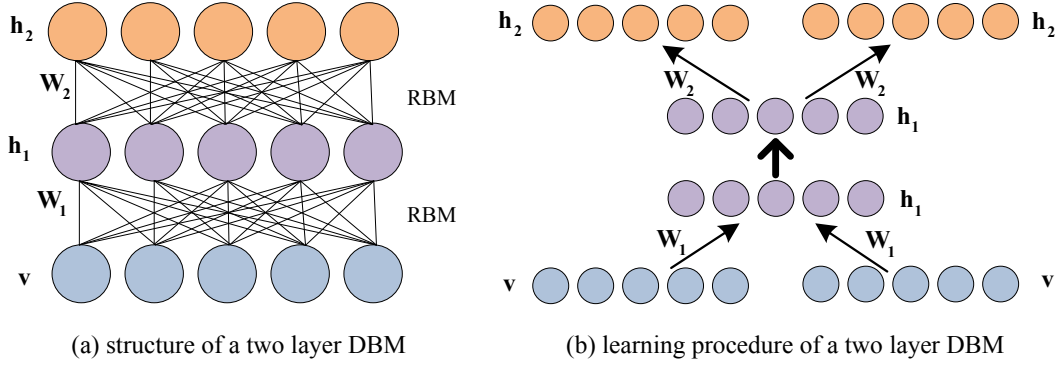

(a) structure of a two layer DBM  (b) learning procedure of a two layer DBM

**Fig. 3 The DBM structure and its learning procedure.** (a) the structure of a two layer DBM where each layer is an RBM, $\mathbf{v}$ stands for visible units, $\mathbf{h}_1$ and $\mathbf{h}_2$ represent the first hidden layer and the second hidden layer respectively. $\mathbf{W}_1$ denotes the weight matrix between the visible layer and the first hidden layer. $\mathbf{W}_2$ stands for the weight matrix between the first hidden layer and the second hidden layer. (b) the learning procedure of the DBM.

The approximate inference procedure of $\mathbf{W}$ for the DBM includes an initial bottom-up pass and a top-down feedback to make the DBM receive back-propagation error from the output layer. As shown in Fig. 3, the DBM model is stacked by RBMs and it has the potential of learning hidden relationships among hidden units even though they become increasingly complicated. In addition, the DBM model can extract features from a large number of unlabeled input data, and then, be fine-tuned by using a few labeled data. Fig. 3(a) illustrates an example of a two-layer DBM model. The energy of the state $\{\mathbf{v},\mathbf{h}_1,\mathbf{h}_2\}$ is defined as

$$E(\mathbf{v},\mathbf{h}_1,\mathbf{h}_2;\boldsymbol{\theta}) = -\mathbf{v}^T\mathbf{W}_1\mathbf{h}_1 - \mathbf{h}_1^T\mathbf{W}_2\mathbf{h}_2 \qquad (8)$$

where $\boldsymbol{\theta}=\{\mathbf{W}_1,\mathbf{W}_2\}$ are the parameters of the model, and the biases of each layer are ignored here. $\mathbf{W}_1$ represents the connection matrix between the visible layer and the hidden layer, and $\mathbf{W}_2$ denotes the connection matrix between the first hidden layer and the second hidden layer. The probability over the visible units $\mathbf{v}$ assigned by the

model is:

$$p(\mathbf{v};\boldsymbol{\theta}) = \frac{1}{Z(\boldsymbol{\theta})} \sum_{\mathbf{h_1h_2}} \exp(-E(\mathbf{v},\mathbf{h_1},\mathbf{h_2};\boldsymbol{\theta})) \qquad (9)$$

The conditional distributions between the visible and the hidden units are:

$$p(h_{1j}=1|\mathbf{v},\mathbf{h_2}) = \sigma\left(\sum_i W_{1ij}v_i + \sum_m W_{2jm}h_{2j}\right) \qquad (10)$$

$$p(h_{2m}=1|\mathbf{h_1}) = \sigma\left(\sum_i W_{2im}h_{1i}\right) \qquad (11)$$

$$p(v_i=1|\mathbf{h_1}) = \sigma\left(\sum_j W_{1ij}h_{1j}\right) \qquad (12)$$

Concerning the MLE learning, we can still adapt the learning procedure of standard Boltzmann machines, but it runs rather slowly, especially when the amount of hidden layers is large. Therefore, the greedy layer-wise algorithm [7] is a better way to learn the DBM model, which can quickly initialize the model parameters to suited values.

**Learning with Greedy Layer-wise**

Greedy layer-wise, as the name implies, is an unsupervised layer-by-layer learning algorithm. As shown in Fig. 3(a), two adjacent layers are regarded as an RBM, and each layer is trained in the way as described in section 2.2. The procedure of greedy layer-wise learning is: first, train the first RBM which consists of a visible layer $\mathbf{v}$ and the first hidden layer $\mathbf{h_1}$. Then, compute $p(\mathbf{h_1}|\mathbf{v})$ (see equation (4)) using the trained $\mathbf{W_1}$ and substitute for $\mathbf{h_1}$. Next, train the second RBM as described in section 2.2, regarding the updated $\mathbf{h_1}$ and $\mathbf{h_2}$ as the visible layer and the hidden layer, respectively. Other layers of the DBM model can also be trained in the same way as mentioned above.

As shown in Fig. 3(b), when training the DBM model, the value of $\mathbf{h}_1$ is not only related to $\mathbf{W}_1$ but also to $\mathbf{W}_2$. Thus, $\mathbf{h}_1$ is calculated using $\mathbf{W}_1/2$ bottom-up and $\mathbf{W}_2/2$ top-down, which combines both top-down and bottom-up influence. In order to conveniently compute this, we adapt the approach of doubling the number of units of visible layers and top hidden layers, and tie the visible-to-hidden weights of the first layer RBM and the top-level RBM to two copies, as described in [10].

The conditional distributions of the hidden and visible states take the form

$$p(h_{1j}=1|\mathbf{v}) = \sigma\left(\sum_i W_{1ij}v_i + \sum_i W_{1ij}v_i\right) \tag{13}$$

$$p(v_i=\mathbf{1}|\mathbf{h}_1) = \sigma\left(\sum_j W_{1ij}h_{1j}\right) \tag{14}$$

where the CD algorithm and Gibbs sampling, as described in section 2.2, are employed to update $W_{ij}$.

For the second RBM, the first hidden layer $\mathbf{h}_1$ is regarded as the visible layer and its value equals to $p(\mathbf{h}_1;\mathbf{W}_1)$; and $\mathbf{h}_2$ is regarded as the hidden layer. Since $\mathbf{h}_2$ is also the top hidden layer in the two layer DBM, the rule of doubling units and weights mentioned above is also applied to $\mathbf{h}_2$. Thus, the conditional distributions for the second RBM are defined as

$$p(h_{1j}=1|\mathbf{h}_2) = \sigma\left(\sum_m W_{jm}^2 h_{2m} + \sum_m W_{jm}^2 h_{2m}\right) \tag{15}$$

$$p(h_{2m} = 1|\mathbf{h_1}) = \sigma\left(\sum_j W_{jm}^2 h_{1j}\right) \tag{16}$$

After training the adjacent stacked RBMs, the value of $\mathbf{h_1}$ would be recomputed using the neighboring RBMs with the trained $\mathbf{W_1}$ and $\mathbf{W_2}$. So $\mathbf{h_1}$ is redefined by $\mathbf{v}$ and $\mathbf{h_2}$ as

$$p(h_{1j} = 1|\mathbf{v}, \mathbf{h_2}) = \sigma\left(\sum_i W_{1ij} v_i + \sum_m W_{2jm} h_{2m}\right) \tag{17}$$

where $v_i$ and $h_{2m}$ are the same as in equations (14) and (16) respectively.

## Preprocessing with the SLIC

As shown in Fig. 3, one unit of the visual layer in a DBM model represents one pixel. If the input image is large, the computational complexity of learning a DBM model will increase rapidly. Therefore, it is necessary to perform an image preprocessing to reduce feature dimensionality.

Generally speaking, convolution and pooling are two effective ways to reduce the image size [13]. Using the pooling algorithm to resize a large natural image, however, will lose much more information; and the pooling algorithm is limited by the coordinate of pixels. In addition, utilizing a convolution algorithm for preprocessing large images is a time-consuming process. So it is of great importance to find a suitable preprocessing approach for large image recognition. Superpixels have become a popular preprocessing method in computer vision applications, such as segmentation and object localization.

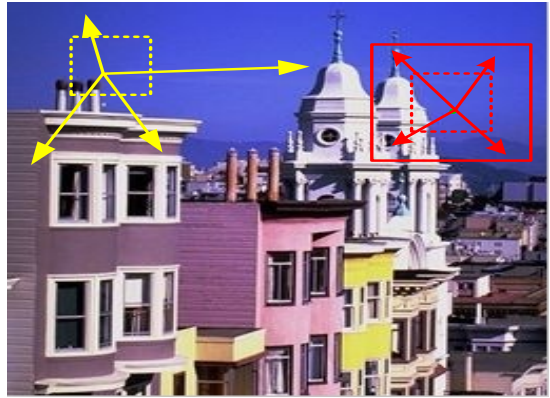

**Fig. 4 Illustration of searching region for a new cluster center.** Yellow arrows are the searching region of standard superpixels in the whole image. Red arrows represent SLIC searching region which becomes $2S \times 2S$.

## Superpixels

A superpixel is a meaningful atomic region combining similar pixels according to the texture, color, location, etc. superpixel can reduce the image redundancy and greatly decrease the complexity of subsequent image processing tasks [18]. In recent years, many superpixel algorithms have been proposed for image segmentation [19].

In general, superpixel algorithms can be classified into two categories: graph-based methods and gradient ascent methods. Graph-based approaches treat each pixel as a node and the relationship between neighboring nodes as an edge, similar to graph models. The edges are computed by minimizing a cost function defined by prior knowledge. Gradient ascent methods start from a random initial gathering of pixels, and then iteratively compute the clusters until convergence.

## Simple Linear Iterative Clustering

Simple linear iterative clustering (SLIC) is proposed by *Achanta et al.*[18]*,* which is faster and more memory efficient than other superpixel methods. Since there is onlyone parameter $k$, the desired number of superpixels, the SLIC is easy to implement and understand. Firstly, a cluster center, $k=1,2,...,K$ is randomly chosen in an $S \times S$ region, where $S=\sqrt{N/K}$ is the region size and $N$ is the total number of pixels in an image. Find the lowest gradient pixel to replace $C_k$ in the $3 \times 3$ neighborhood. Then, in order to decrease computational complexity, a local *k*-means algorithm is employed to assign each pixel around $C_k$ neighboring $2S \times 2S$ region to the nearest cluster center as shown in Fig. 4. The cluster center $C_k$ and the pixel $i$ label $lab(i)$ are updated according to the distance $D(i)$ between $C_k$ and pixel $i$. Since color and position in space are the most obvious information in an RGB image, the distance $D(i)$ is defined by combining color distance with space distance in this paper. The distance between two colors is a metric of interest in color science [20]. For the SLIC, CIE76 [21], the first color distance formula is selected as the color distance that relates a measure to a known Lab (label) value, and is defined as the distance between two colors is a metric of interest in color science [20]. For the SLIC, CIE76 [21], the first color distance formula is selected as the color distance that relates a measure to a known Lab (label) value, and is defined as

$$d_c = \sqrt{(l_j - l_i)^2 + (a_j - a_i)^2 + (b_j - b_i)^2} \tag{18}$$

where $l$ stands for luminance. The parameters $a$ and $b$ respectively represent green or red and blue or yellow. The Euclidean distance is defined as the spatial distance

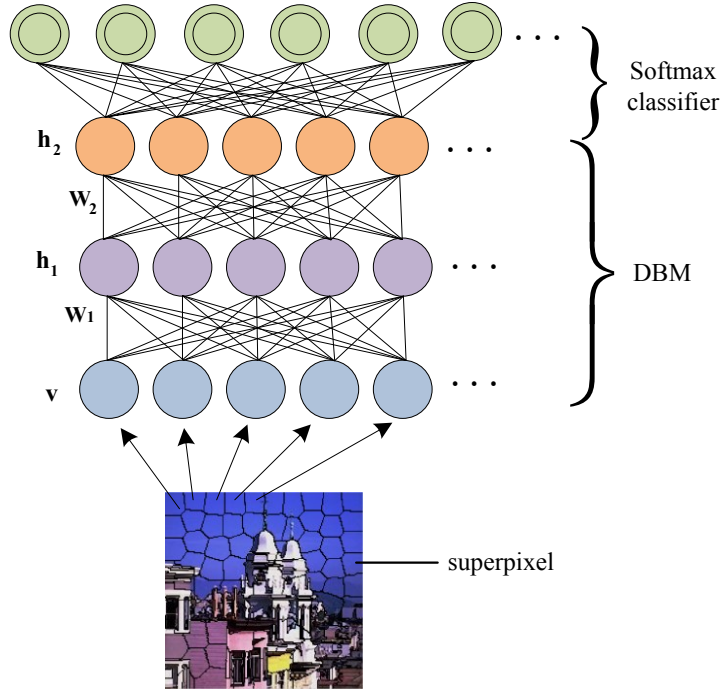

**Fig. 5 Illustration of natural scene recognition based on the DBM model.** The DBM is used to extract features after preprocessing of superpixels, while the softmax classifier is employed to classify extracted features.

$$d_s = \sqrt{(x_j - x_i)^2 + (y_j - y_i)^2} \quad (19)$$

Thus, the distance $D$ is defined as:

$$D = \sqrt{\left(\frac{d_c}{N_c}\right)^2 + \left(\frac{d_s}{N_s}\right)^2} \quad (20)$$

where $N_S = S = \sqrt{(N/K)}$ and $N_c$ is determined by the signification of the color distance between two clusters. In this paper, we fix $N_c$ to a constant value $m$. After simplification, $D$ is written as

$$D = \sqrt{d_c + \left(\frac{d_s}{S}\right)^2 m^2} \quad (21)$$

Finally, the $L_1$ norm is adopted to represent the residual error $E$ between the new cluster center and the previous one. We then iteratively repeat the update steps until $E$ converges. In order to enforce connectivity, the disjoint pixels are reassigned to its nearby superpixels.

## Dimensional Reduction by the SLIC

The SLIC is widely used in image segmentation due to its superiority in computational speed. This algorithm substantially integrates the pixels based on the similarity in location and color. This idea can also be used to reduce image dimensions. For example, after generating superpixels, each superpixel can be regarded as a new pixel in the image, which can effectively reduce image dimensions. Different to convolution and pooling methods that only rely on pixel location, the SLIC combines location and color information to enhance the performance of dimensional reduction, which is helpful for subsequent image processing tasks.

## Scene Recognition with Softmax Regression

In order to perform scene recognition, softmax regression was used to classify the scene images in our experiments. The architecture of the proposed model for natural scene images is shown in Fig. 5. Softmax regression is a variant of the logistic regression model which is only applied to binary classification. In this paper, we are interested in multi-class classification. When setting the softmax classifier, the label can take on different values rather than only two. Thus, we have the training set $\{(x^{(1)}, y^{(1)}), ..., (x^{(m)}, y^{(m)})\}$ of $m$ labeled examples and $y^{(i)} \in \{1, 2, 3, ..., k\}$. To obtain the probability of the class labels from the $k$ possible values given a test input $x$, the hypothesis is estimated by the probability $p(y=j|x)$ for each value of $j=1,2,...,k$. Therefore, the hypothesis will yield a $k$-dimensional vector (whose elements sum to 1) giving $k$ estimated probabilities. Specifically, the hypothesis function $h_{\theta(x)}$ takes the following form.

$$h_\theta(x^{(i)}) = \begin{bmatrix} p(y^{(i)}=1|x^{(i)};\theta) \\ p(y^{(i)}=2|x^{(i)};\theta) \\ p(y^{(i)}=3|x^{(i)};\theta) \\ \vdots \\ p(y^{(i)}=k|x^{(i)};\theta) \end{bmatrix} = \frac{1}{\sum_{j=1}^{k} e^{\theta_j^T x(i)}} \begin{bmatrix} e^{\theta_1^T x(i)} \\ e^{\theta_2^T x(i)} \\ e^{\theta_3^T x(i)} \\ \vdots \\ e^{\theta_k^T x(i)} \end{bmatrix} \quad (22)$$

where $\theta$ is the parameter of the softmax regression model; and $\sum_{j=1}^{k} e^{\theta_j^T x(i)}$ is a normalization term for the distribution.

The parameter $\theta$ is learned in order to minimize the cost function. In equation (23), $1\{\cdot\}$ is the indicator function, that is $1\{a\ true\ statement\}=1$, and $1\{a\ false\ statement\}=0$. For example, $1\{1+1=2\}$ evaluates to 1; and $1\{1+1=4\}$ evaluates to 0. So the cost function is written as

$$\begin{aligned} J(\theta) &= -\frac{1}{m}\left[\sum_{i=1}^{m}(1-y^{(i)})\log(1-h_\theta(x^{(i)})) + y^{(i)}\log h_\theta(x^{(i)})\right] \\ &= -\frac{1}{m}\left[\sum_{i=1}^{m}\sum_{j=0}^{1} 1\{y^{(i)}=j\}\log p(y^{(i)}=1|x^{(i)};\ \theta)\right] \end{aligned} \quad (23)$$

where θ is initialized as any random number. Note that in softmax regression, $p(y^{(i)}=k|x^{(i)};\boldsymbol{\theta})$ is known as

$$p\left(y^{(i)} = j \middle| x^{(i)}; \boldsymbol{\theta}\right) = \frac{e^{\theta_j^T x(i)}}{\sum_{l=1}^{k} e^{\theta_l^T x(i)}} \tag{24}$$

There is no closed-form approach to solve for the minimum of $J(\theta)$, and thus, some iterative optimization algorithm, such as gradient descent or L-BFGS(Limited-memory BFGS), can be adopted. Taking derivative of the cost function (24), we have

$$\nabla_{\theta_j} J(\theta) = -\frac{1}{m} \sum_{i=1}^{m} \left[ x^{(i)} \left( 1\{y^{(i)} = j\} - p\left(y^{(i)} = 1 \middle| x^{(i)}; \theta\right) \right) \right] \tag{25}$$

where, $\nabla_{\theta_j}$ is a vector whose *l-th* element is the partial derivative of $J(\theta)$ with respect to the *l-th* element of $\theta_j$. Then, in each iterations, $\theta_j$ can be updated by $\theta_j = \theta_j - \alpha \nabla_{\theta_j} J(\theta)$, with $j = 1, \ldots, k$.

When implementing softmax regression, we usually use a modified version of the cost function by adding a decay term as below. The weight decay term avoids the solution of large values of the parameters.

$$J(\theta) = -\frac{1}{m} \left[ \sum_{i=1}^{m} \sum_{j=0}^{1} 1\{y^{(i)} = j\} \log p\left(y^{(i)} = 1 \middle| x^{(i)}; \boldsymbol{\theta}\right) \right] + \frac{\lambda}{2} \sum_{i=1}^{k} \sum_{j=0}^{n} \theta_{ij}^2 \tag{26}$$

The implementation of the proposed algorithm based on superpixels and the DBM for natural scene images is summarized as follows.

(1) Generate superpixels using SLIC.

    a) Randomly select an initial cluster centers in each $S \times S$ region, and find the lowest gradient pixels to replace the cluster centers in a $3 \times 3$ neighbor region;

b) For each cluster center, calculate the distance $D$ between the cluster center and each pixel in the $2S \times 2S$ neighbor region, and update the cluster centers ;

c) Calculate the residual error $E$ using $L1$ norm distance, repeat step b) until $E < threshold$ ;

(2) Take the superpixels obtained in (1) as the input data of the visible units, train the DBM model using the CD algorithm and the Gibbs sampling.

a) Given the visible units $\mathbf{v}^1$, calculate $p(\mathbf{h}_1^1|\mathbf{v}^1)$, and replace $\mathbf{h}_1$ by the value of $p(\mathbf{h}_1^1|\mathbf{v}^1)$;

b) Compute $p(\mathbf{v}^2|\mathbf{h}_1^1)$, and set the reconstructed value $\mathbf{v}^2$ as $p(\mathbf{v}^2|\mathbf{h}_1^1)$;

c) Use the Gibbs sampling to get $\mathbf{v}^n$ and $\mathbf{h}_1^n$; and update the weight matrix $\mathbf{W}_1$ using the CD algorithm;

d) Repeat steps a) to c) iteratively until convergence or reaching the maximum number of iterations;

e) Repeat steps a) to d) to compute the second layer and learn the weight matrix $\mathbf{W}_2$;

f) Rconstruct $\mathbf{h}_1$ using $\mathbf{W}_1$ and $\mathbf{W}_2$ according to equation (17);

(3) Repeat step (2) to train other RBMs in the DBM.

(4) Employ softmax regression to categorize scene images using the extracted features.

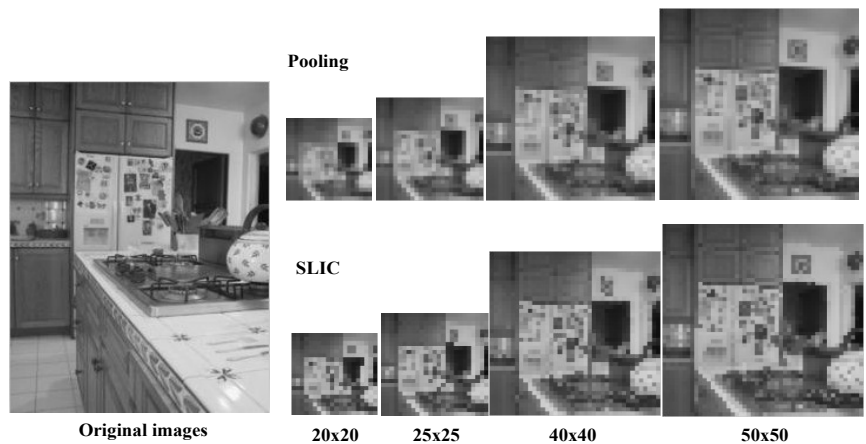

**Fig. 6 The preprocessed images with different sizes pooling and SLIC respectively.** The top row images are preprocessed using pooling and the bottom images are preprocessed by SLIC. The papers on the wall in the bottom row are more distinct than the ones in the top row, which breaks the specific shape of the computational region.

# Experiments and Results

We select three different public scene datasets to evaluate the proposed approach. In this paper, all experiments were performed on a Sony PC with an Intel Core i3 CPU 350 @2.27GHz, and 6GB of random access memory. The experimental results are reported as below.

## Fifteen-Scene Dataset

The fifteen-Scene dataset [22] includes 4485 images of fifteen-scene categories describing different indoor and outdoor scenes. The size of the images is roughly $200 \times 300$ pixels. In this paper, we respectively used the SLIC and the pooling approach to preprocess images. First, the all images are resized to $200 \times 200$ pixels. Then, the SLIC and pooling were respectively applied to reduce the image sizes to $20 \times 20$ pixels, $25 \times 25$ pixels, and $40 \times 40$ pixels. The results are shown in Fig. 6, from which we can see that the SLIC can catch more boundary information of the objects. In Fig. 6, the papers on the wall can be distinguished in images ranging from $25\times25$ pixels to $50\times50$ pixels preprocessed by the SLIC, while they can only be identified in the $50\times50$ pixel images using the pooling algorithm. We also preprocessed different categories of scene images, both indoor and outdoor, using the SLIC and pooling. Some example images and the corresponding preprocessed results of $40\times40$ pixels are shown in Fig. 7. It can be seen that the results of the SLIC are more distinct and the curvatures of objects are clearly visible, especially in the third-column and the forth-column.

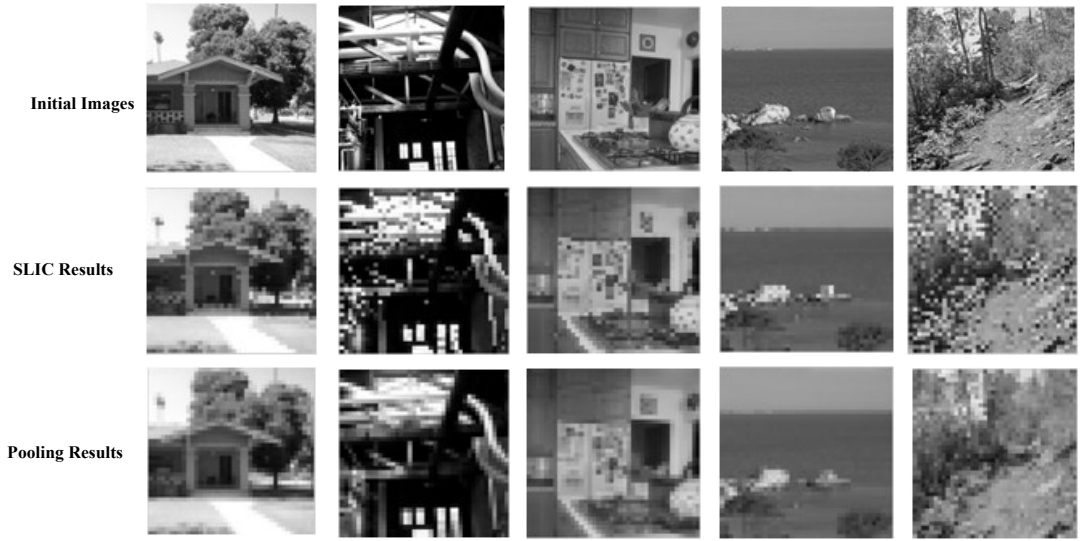

**Fig. 7 The preprocessing results of scene images using the pooling and the SLIC, respectively.** The original image sizes are all 200x200 pixels and the preprocessed image sizes are all 40x40 pixels. The SLIC results are more distinct, and the curvature of objects was clearly visible, especially in the third and fourth columns.

**Table 1.** Recognition rate of different sizes of preprocessed images

| Preprocessing Method \ Image Size | 20x20 pixels | 25x25 pixels | 40x40 pixels | 50x50 pixels |
|---|---|---|---|---|
| **Pooling** | 81.3% | 84.7% | 87.2% | 88.4% |
| **SLIC** | 80.9% | 89.7% | 93.2% | 93.8% |

**Table 2**. Recognition rate over 15-scene dataset

| Method | Recognition Rate(%) |
|---|---|
| GIST-color[26] | 69.5 |
| RBow[27] | 78.60 |
| Classmes[28] | 80.60 |
| Object Bank[29] | 80.90 |
| SP[16] | 81.40 |
| SPMSM[30] | 82.30 |
| LCSR[31] | 82.67 |
| SP-pLSA[32] | 83.70 |
| CENTRIST[33] | 83.88 |
| HIK[34] | 84.12 |
| VC+VQ[35] | 85.40 |
| LMLF[36] | 85.60 |
| LPR[4] | 85.81 |
| RSP[3] | 88.10 |
| LScSPM[37] | 89.75 |
| ISPR+IFV[5] | 91.06 |
| **Our Approach** | **93.50** |

In the experiments, we selected 200 images per class for training and 20 images per class for testing. We constructed the DBM stacked by two RBMs to extract image features. Since the input images are preprocessed into different sizes, we constructed different structural DBMs for scene recognition. For the images of $20 \times 20$ pixels and $25 \times 25$ pixels, we built a DBM with 300 units of the first hidden layer and 200 units of the second hidden layer. For larger input images of $40 \times 40$ pixels and $50 \times 50$ pixels, we constructed a DBM with 1000 units in the first hidden layer and 500 units in the second hidden layer to extract features. Table.1 shows the recognition results using different sizes of preprocessed images as the input data for the fifteen-scene dataset. As demonstrated in Table.1, the recognition rate is improved with the increase of the input image size; in every cases, the results of the SLIC are better than those from the pooling method. For example, the recognition rate of the input images of $50 \times 50$ pixels is the highest in our experimental results. However, it is worth noting that the computing cost will increase dramatically with the increase of image sizes. When the input image size is $50 \times 50$ pixels, the real computation time is 34775 seconds for recognizing the fifteen-scene dataset, while the consuming time is 4277 seconds when the input image size is $40 \times 40$ pixels.

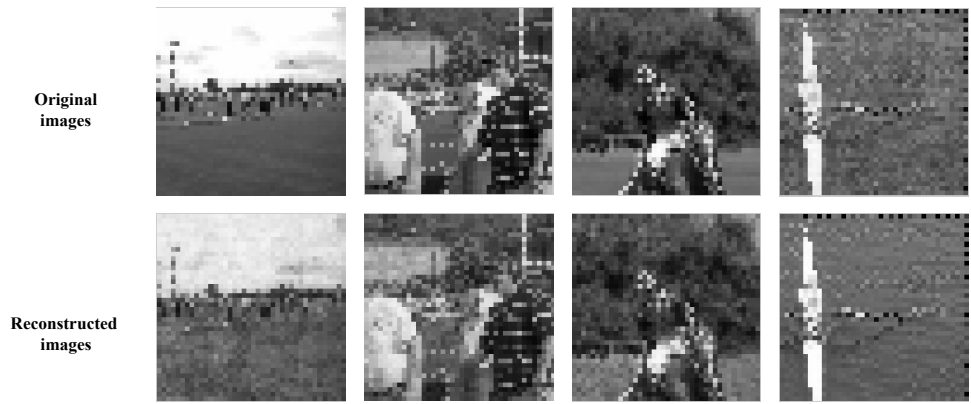

**Fig. 8 Some reconstruction images using the UIUC 8-sports dataset.** The top row shows the original images preprocessed by the SLIC; and the bottom row shows the reconstructed images. The reconstructed images contain most of the information from the original images.

Table.2 shows the recognition rates of our approach compared to other methods, where the DBM model was built with 1000 units for the first hidden layer and 500 units for the second hidden layer for the input images of $40 \times 40$ pixels. It is evident that the proposed approach outperforms all other counterparts, including the most recent technique [5] published at CVPR 2014, in terms of accuracy for natural scene image recognition.

## UIUC 8-Sports Dataset

The UIUC 8-sports dataset was built by *Li-Jia* and *Fei-Fei* [23]. It contains eightdifferent sport scenes: rowing badminton, polo, bocce, snowboarding, croquet, sailing, and rock climbing. The image sizes are around 1000×800 pixels. In our experiments, we first resized all images to 600×600 pixels; and then, their dimensionality is further reduced to 50×50 pixels using the SLIC. We selected 100 images per class as the training set and 20 images per class as the testing set. The structure of the DBM contains 1000 units for the first hidden layer and 500 units for the second hidden layer. Some original images sampled from the training set and their reconstructed images are showed in Fig. 8. We can see that the reconstructed images using the extracted features contain most of the information from the original images.

Table 3. Recognition rate over UIUC8-sports dataset

| Method | Recognition Rate(%) |
|---|---|
| GIST-color[26] | 70.70 |
| RBow[27] | 71.70 |
| Classmes[28] | 73.4 |
| Object Bank[29] | 76.3 |
| SP[16] | 79.52 |
| SPMSM[30] | 79.6 |
| LCSR[31] | 81.80 |
| SP-pLSA[32] | 83.00 |
| CENTRIST[33] | 84.20 |
| HIK[34] | 85.2 |
| VC+VQ[35] | 85.3 |
| LMLF[36] | 86.25 |
| LPR[4] | 88.37 |
| RSP[3] | 88.40 |
| LScSPM[37] | 90.92 |
| ISPR+IFV[5] | 92.31 |
| **Our Approach** | **92.50** |

After extracting the features, the softmax regression is applied to categorize scenes over the UIUC 8-sports dataset. As shown in Table.3, the recognition rate of the proposed method is higher than all state-of-the-art approaches.

**SIFT Flow Dataset**

The SIFT Flow dataset [24] consists of 2,688 images and is split into 2,488 training images and 200 test images. The dataset was employed in scene labeling and scene parsing[25]. The size of all images is 256x256 pixels. *Cle´ment Farabet* [25] employed the convolutional network (ConvNet), which is another popular model of deep learning, to recognize scenes on the SIFT Flow dataset. In this paper, we also test the performance of our method using this dataset. First, we reduced the size of the images to $32\times32$ pixels. Then, we trained a DBM with 1024 visible units, 500 units of the first hidden layer, and 200 units of the second hidden layer. Finally, the extracted features were classified by using the softmax classifier. The recognition rate for the SIFT Flow dataset achieves 80.1%.

**Discussions**

Deep learning models, such as the DBN and the DBM, have attracted more and more attention, and been successfully applied to the recognition tasks. However, since the sizes of natural images are always very large, the problem of computational complexity must be considered when designing the DBM model for scene recognition. In this paper, we presented a scene recognition method that combines superpixels and the DBM model. Since the SLIC can generate superpixels efficiently, our method performs better than the pooling algorithm for dimensionality reduction of natural images. The experimental results on the fifteen scene dataset, UIUC 8-sports dataset, and the SIFT Flow dataset show that the proposed method can obtain the best performance than other counterparts in terms of recognition rate. During the experiments, however, we also find that the structure of the constructed DBM model will influence the recognition results. For input images with $40 \times 40$ pixels, the DBM with 400 units of the first hidden layer and 200 units of the second hidden layer obtained a recognition rate of 73.7%, as a contrast, the DBM with 1000 units of the first hidden layer and 500 units of the second hidden layer achieved a recognition rate 93.2%. In addition, using the DBM model to extract features is computational intensive when the input data size becomes large. For example, when we use a two-layer DBM with 1000 first hidden units and 500 second hidden units to extract features for the fifteen-scene dataset, the real computation time decreases from 34775 to 4277 seconds if the input images are decreased from $50 \times 50$ to $40 \times 40$ pixels.

## Conclusions

In this paper, we have presented a new approach for scene recognition based on superpixels and the DBM model. First, we used the SLIC to preprocess large natural images, which can effectively reduce the computational complexity for subsequent image processing tasks. Compared to the pooling algorithm, the SLIC preprocessing can preserve more information found in the images, which is critical for scene recognition. Then, we constructed a two-layer DBM model to extract features in an unsupervised manner and utilized the softmax regression to classify the scenes. The experimental results over the fifteen-scene dataset, the UIUC 8-Sports Dataset, and the SIFT Flow dataset demonstrate that the proposed method performs better than other counterparts in terms of scene recognition accuracy. In the future study, we will further investigate how to construct a more effective structure of the DBM model for practical applications.

## Acknowledgement

This work is partly supported by the National Natural Science Foundation of China under grant Nos. 61201362 and 61273282, the Beijing Natural Science Foundation under grant No.7132021, and the Scientific Research Project of Beijing Educational Committee under grant no. KM201410005005.